\documentclass[11pt,letterpaper]{article}
\usepackage{emnlp2017}
\usepackage{times}
\usepackage{latexsym}

\emnlpfinalcopy

\def\emnlppaperid{1017}

\usepackage{url}
\usepackage[normalem]{ulem}	
\usepackage{multirow}
\usepackage[mathscr]{euscript}
\usepackage{graphicx}
\usepackage{enumitem}
\usepackage{subfigure}
\usepackage{caption}
\usepackage{color}
\usepackage{amsmath,amsfonts,amssymb}

\newcommand{\ignore}[1]{}

\title{No Need to \emph{Pay Attention}:\\ Simple Recurrent Neural Networks Work!\\(for Answering ``Simple'' Questions)}

\author{Ferhan Ture \and Oliver Jojic\\
  Comcast Labs, Washington, DC 20005\\
  {\tt {\small \{ferhan\_ture, oliver\_jojic\}@cable.comcast.com}}}
  
\date{}

\begin{document}

\maketitle

\begin{abstract}
First-order factoid question answering assumes that the question can be answered by a single
fact in a knowledge base (KB). While this does not seem like a challenging task, many recent attempts that apply 
either complex linguistic reasoning or deep neural networks achieve 65\%--76\% accuracy on benchmark
sets. Our approach formulates the task as two machine learning problems:\ detecting the entities
in the question, and classifying the question as one of the relation types in the KB.
We train a recurrent neural network to solve each problem.
On the SimpleQuestions dataset, our approach yields substantial improvements over previously published results ---
even neural networks based on much more complex architectures. The simplicity of our approach
also has practical advantages, such as efficiency and modularity, that are valuable especially in an industry setting.
In fact, we present a preliminary analysis of the performance of our model on real queries from Comcast's X1
entertainment platform with millions of users every day.
\end{abstract}

\section{Introduction}\label{sec:intro}
First-order factoid question answering (QA) assumes that the question can be answered 
by a single fact in a knowledge base (KB). For example, ``How old is Tom Hanks'' is about the [age] of 
[Tom Hanks].
Also referred to as \emph{simple} questions by Bordes et al.~\shortcite{Bordes:2015aa}, recent attempts that apply 
either complex linguistic reasoning or attention-based complex neural network architectures achieve up to 76\% accuracy 
on benchmark sets~\cite{Golub:2016aa,Yin:2016aa}. While it is tempting to study QA systems that can handle more 
complicated questions, it is hard to reach reasonably high precision for unrestricted questions.
For more than a decade, successful industry applications of QA have focused on first-order 
questions.
This bears the question:\ are users even interested in asking questions beyond first-order (or are these use cases 
more suitable for interactive dialogue)? Based on voice logs from a major entertainment platform with millions of 
users every day, Comcast X1, we find that most existing use cases of QA fall into the first-order category.


Our strategy is to tailor our approach to first-order QA by making strong assumptions about the problem structure. 
In particular, we assume that the answer to a first-order question is a \emph{single} property of 
a \emph{single} entity in the KB, and decompose the task into two subproblems:\ (a) detecting entities
in the question and (b) classifying the question as one of the relation types in the KB.
We simply train a vanilla recurrent neural network (RNN) to solve each subproblem~\cite{Elman:1990aa}.
Despite its simplicity, our approach (RNN-QA) achieves the highest reported accuracy on the SimpleQuestions dataset. 
While recent literature 
has focused on building more complex neural network architectures with attention mechanisms, attempting to 
generalize to broader QA, we enforce stricter assumptions on the problem structure, thereby 
reducing complexity.
This also means that our solution is efficient, another critical requirement for real-time QA applications. 
In fact, we present a performance analysis of RNN-QA on Comcast's X1 entertainment system, used by millions of
customers every day.

\section{Related work}

If knowledge is presented in a structured form (e.g., knowledge base (KB)), the standard approach to QA 
is to transform the question and knowledge into a compatible form, and perform reasoning to determine
which fact in the KB answers a given question. Examples of this approach include 
pattern-based question analyzers~\cite{Buscaldi:2010aa}, combination of syntactic parsing
and semantic role labeling~\cite{Bilotti:2007aa,Bilotti:2010aa}, as well as lambda calculus~\cite{Berant:2013aa} and combinatory 
categorical grammars (CCG)~\cite{Reddy:2014aa}. 
A downside of these approaches is the reliance on linguistic resources/heuristics, making them 
language- and/or domain-specific. Even though Reddy et 
al.~\shortcite{Reddy:2014aa} claim that their approach requires less supervision than prior work, it still relies on many 
English-specific heuristics and hand-crafted features. Also, their most accurate model uses a corpus of paraphrases
to generalize to linguistic diversity. Linguistic parsers can also be too slow for real-time applications.

In contrast, an RNN can detect 
entities in the question with high accuracy and low latency. The only required resources are word embeddings and a set of 
questions with entity words tagged. The former can be easily trained for any language/domain in an 
unsupervised fashion, given a large text corpus without annotations~\cite{Mikolov:2013aa,Pennington:2014aa}. 
The latter is a relatively simple annotation task that exists for many languages and domains, and it can also be synthetically 
generated. 
Many researchers have explored similar techniques for general NLP tasks~\cite{Collobert:2011aa}, such as named entity 
recognition~\cite{Lu:2015aa,Hammerton:2003aa}, sequence labeling~\cite{Graves:2008aa,Chung:2014aa}, part-of-speech 
tagging~\cite{Huang:2015aa,Wang:2015aa}, chunking~\cite{Huang:2015aa}.

Deep learning techniques have been studied extensively for constructing parallel neural networks for modeling a joint 
probability distribution for question-answer pairs~\cite{Hsu:2016aa,Yang:2014aa,He:2015aa,Mueller:2016aa} and 
re-ranking answers output by a retrieval engine~\cite{Rao:2016aa,Yang:2016aa}.
These more complex approaches might be needed for general-purpose QA and sentence similarity, where one cannot make 
assumptions about the structure of the input or knowledge. However, as noted in Section~\ref{sec:intro}, first-order 
factoid questions can be represented by an entity and a relation type, and the answer is usually stored in a structured
knowledge base. Dong et al.~\shortcite{Dong:2015ab} similarly assume that the answer to a question is at most two
hops away from the target entity. However, they do not propose how 
to obtain the target entity, since it is provided as part of their dataset. Bordes et al.~\shortcite{Bordes:2014aa} 
take advantage of the KB structure by projecting entities, relations, and subgraphs into the same latent space.
In addition to finding the target entity, the other key information to first-order QA is the relation
type corresponding to the question. Many researchers have shown that classifying the question into one of the 
pre-defined types (e.g., based on patterns~\cite{Zhang:2003aa} or support vector 
machines~\cite{Buscaldi:2010aa}) improves QA accuracy.

\section{Approach}\label{sec:approach}


\smallskip \noindent {\bf (a) From Question to Structured Query.}
Our approach relies on a \emph{knowledge base}, containing a large set of \emph{facts}, each one 
representing a binary [subject,~relation,~object] relationship. 
Since we assume \emph{first-order questions}, the answer can be retrieved from a single fact.
For instance, ``How old is Sarah Michelle Gellar?'' can be answered by the fact:
\vspace{-0.15cm}
\begin{flushleft}
{\footnotesize 
\verb|[Sarah Michelle Gellar,bornOn,4/14/1977]   |
\vspace{0.3cm}
}
\end{flushleft}


The main idea is to dissect a first-order factoid natural-language question by converting it into 
a structured query:
\{\textrm{entity ``Sarah Michelle Gellar'', relation}: \verb|bornOn|\}.
The process can be modularized into two machine learning problems, namely \emph{\textbf{entity detection}} and 
\emph{\textbf{relation prediction}}. 
In the former, the objective is to \emph{tag} each question word as either entity or not.
In the latter, the objective is to \emph{classify} the question into one of the $K$ relation types.
We modeled both using an RNN.

We use a standard RNN architecture:\ Each word in the question passes through an embedding lookup layer $E$, 
projecting the one-hot vector into a $d$-dimensional vector $x_t$. A recurrent layer combines this \emph{input representation} 
with the hidden layer representation from the \emph{previous word} and applies a non-linear transformation to compute 
the hidden layer representation for the \emph{current word}. The hidden representation of the final recurrent layer is 
projected to the output space of $k$ dimensions and normalized into a probability distribution via soft-max.
\ignore{
\begin{eqnarray}
\ignore{\nonumber x_t=E_{x_t} & \textrm{word repr. at $t$} \\}
\nonumber h_t=\textrm{ReLu}(W h_{t-1}+U x_t) & \textrm{hidden repr. at $t$}\\
\nonumber o_t=\textrm{softmax}(V h_t) & \textrm{output at $t$}
\nonumber \\
\nonumber E=|\textrm{vocab size}| \times d & \textrm{ embedding}\\
\nonumber W=m\times m &  \textrm{transition weights}\\
\nonumber U=m\times d & \textrm{input weights}\\
\nonumber V=k\times d & \textrm{output weights}
\end{eqnarray}
}

In \emph{relation prediction}, the question is classified into one of the 1837 classes (i.e., relation types in Freebase).
In the \emph{entity detection} task, each word is classified as either \emph{entity} or \emph{context} (i.e., $k=2$). 

Given a new question, we run the two RNN models to construct the 
structured query.
Once every question word is classified as entity (denoted by \texttt{E}) or context (denoted by \texttt{C}), 
we can extract entity phrase(s) by grouping consecutive entity words.
For example, for question ``How old is Michelle Gellar'', the output of entity detection is [\texttt{C C C E E}], 
from which we can extract a single entity ``Michelle Gellar''. The output of relation prediction is \texttt{bornOn}.
The inferred structured query $q$ becomes the following:

\noindent $\{\textrm{\emph{entityText}: } \textrm{``michelle gellar''},\textrm{\emph{relation}: } \texttt{bornOn}\}$
	


\smallskip \noindent {\bf (b) Entity Linking.}
The textual reference to the entity (entityText in $q$) needs to be linked to an actual entity node in our KB.
In order to achieve that, we build an \emph{inverted index} $I_{\textrm{entity}}$ that maps all $n$-grams of an entity ($n\in \{1,2,3\}$)
to the entity's alias text (e.g., name or title), each with an associated $TF$-$IDF$ score. We also map the exact text ($n=\infty$) 
to be able to prioritize exact matches. 

Following our running example, let us demonstrate how we construct $I_{\textrm{entity}}$.
Let us assume there is a node $e_i$ in our KB that refers to the 
actress ``Sarah Michelle Gellar''. The alias of this entity node is the name, which has three unigrams (``sarah'', ``michelle'', ``gellar''), 
two bigrams (``sarah michelle'', ``michelle gellar'') and a single trigram (i.e., the entire name). Each one of these $n$-grams gets indexed 
in $I_{\textrm{entity}}$ with $TF$-$IDF$ weights. Here is how the weights would be computed for unigram ``sarah'' and bigram 
``michelle gellar''
($\Rightarrow$ denotes mapping):
\begin{align*}
&I_{\textrm{entity}}(\textrm{``sarah''}) \Rightarrow \{ \textrm{node}: e_i, \\
& \textrm{score}: TF\textrm{-}IDF(\textrm{``sarah''}, \textrm{``sarah michelle gellar''})\} \\
&I_{\textrm{entity}}(\textrm{``michelle gellar''}) \Rightarrow \{ \textrm{node}: e_i, \\
&\textrm{score}: TF\textrm{-}IDF(\textrm{``michelle gellar''}, \\
& \; \;\; \;\; \;\; \;\; \;\; \;\; \;\; \;\; \;\; \;\; \;\; \;\;  \;\textrm{``sarah michelle gellar''})\}
\end{align*}
This is performed for every $n$-gram ($n\in \{1,2,3,\infty\}$) of every entity node in the KB. Assuming there is an entity node, say $e_j$, 
for the actress ``Sarah Jessica Parker'', we would end up creating a second mapping from unigram ``sarah'':
\begin{align*}
&I_{\textrm{entity}}(\textrm{``sarah''}) \Rightarrow \{ \textrm{node}: e_j,\\
							& \textrm{score}: TF\textrm{-}IDF(\textrm{``sarah''}, \textrm{``sarah jessica parker''})\}
\end{align*}
In other words, ``sarah'' would be linked to both $e_i$ and $e_j$, with corresponding $TF$-$IDF$ weights.

Once the index $I_{\textrm{entity}}$ is built, we can link \emph{entityText} from the structured query (e.g., ``michelle gellar'') 
to the intended entity in the KB (e.g., $e_i$). 
Starting with $n=\infty$, we iterate over $n$-grams of \emph{entityText}  and
query $I_{\textrm{entity}}$, which returns all matching entities in the KB with associated $TF$-$IDF$ relevance 
scores. For each $n$-gram, retrieved entities are appended to the candidate set $C$.
We continue this process with decreasing value of $n$ (i.e., $n\in\{\infty,3,2,1\}$)

Early termination
happens if $C$ is non-empty and $n$ is less than or equal to the number of tokens in \emph{entityText}. The latter
criterion is to avoid cases where we find an exact match but there are also partial matches that might
be more relevant:\ For ``jurassic park'', for $n=\infty$, we get  an exact match to the original movie ``Jurassic Park''. 
But we would also like to retrieve ``Jurassic Park II'' as a candidate entity, which is only possible if we keep processing
until $n=2$.

\smallskip \noindent {\bf (c) Answer Selection.}
Once we have a list of candidate entities $C$, we use each candidate node $e_{\textrm{cand}}$ as a starting 
point to reach candidate answers. 

A \emph{graph reachability index} $I_{\textrm{reach}}$ is built for mapping each entity node $e$ to all 
nodes $e'$ that are reachable, with the associated path $p(e,e')$. For the purpose of the current 
approach, we limit our search to a single hop away, but this index can be easily expanded 
to support a wider search.
\begin{align*}
&I_{reach}(e_i) \Rightarrow\\
&\{\textrm{node:} e_{i_1}, \textrm{text:} \textrm{The Grudge}, \textrm{path:} [\texttt{actedIn}]\} \\
&\{\textrm{node:} e_{i_2}, \textrm{text:} \textrm{4/14/1977}, \textrm{path:} [\texttt{bornOn}]\} \\
&\{\textrm{node:} e_{i_3}, \textrm{text:} \textrm{F. Prinze}, \textrm{path:} [\texttt{marriedTo}]\}
\end{align*}
We use $I_{\textrm{reach}}$ to retrieve all nodes $e^\prime$
that are reachable from $e_{\textrm{cand}}$, where the path from is consistent with the predicted 
relation $r$ (i.e., $r\in p(e_{\textrm{cand}},e^\prime$)). These are added to the candidate answer set $A$.
For example, in the example above, node $e_{i_2}$ would have been added to the answer set $A$, since
the path [\texttt{bornOn}] matches the predicted relation in the structured query.
After repeating this process for each entity in $C$, the highest-scored node in $A$ is our best answer to the 
question.

\section{Experimental Setup}\label{sec:eval}
\smallskip \noindent {\bf Data.}
Evaluation of RNN-QA was carried out on SimpleQuestions, which uses a subset of Freebase containing 
17.8M million facts, 4M unique entities, and 
7523 relation types. Indexes $I_\textrm{entity}$ and $I_\textrm{reach}$ are built based on this knowledge base. 


SimpleQuestions was built by 
\cite{Bordes:2014aa} to serve as a larger and more diverse factoid QA dataset.\footnote{75910/10845/21687 question-answer pairs for 
training/validation/test is an order of magnitude larger than comparable datasets. Vocabulary size is 55K as opposed to around 3K for 
WebQuestions~\cite{Berant:2013aa}.} Freebase facts are sampled in a way to ensure a diverse set of questions, then given to human 
annotators to create questions from, and get labeled with corresponding entity and relation type. There are a total of 1837 unique relation 
types that appear in SimpleQuestions.

\smallskip \noindent {\bf Training.} 
We fixed the embedding layer based on the pre-trained 300-dimensional 
Google News embedding,\footnote{\url{word2vec.googlecode.com}} since the data size is too small for training embeddings.
Out-of-vocabulary words were assigned to a random vector (sampled from uniform distribution).
Parameters were learned via stochastic gradient descent, 
using categorical cross-entropy as objective. In order to handle variable-length input, we limit the input 
to $N$ tokens and prepend a special pad word if input has fewer.\footnote{Input length ($N$) was set to 36, the maximum 
number of tokens across training and validation splits. }
We tried a variety of configurations for the RNN:\ four choices for the type of RNN layer (GRU or LSTM, 
bidirectional or not); depth from 1 to 3; and drop-out ratio from 0 to 0.5, yielding a total of 48 possible configurations. 
For each possible setting, we trained the model on the training portion and used the validation portion to avoid
over-fitting. 
After running all 48 experiments, the most optimal setting was selected by micro-averaged F-score of predicted entities (entity detection)
or accuracy (relation prediction) on the validation set. We concluded that the optimal model is a 2-layer bidirectional LSTM (BiLSTM2) 
for entity detection and BiGRU2 for relation prediction. Drop-out was 10\% in both cases.



%

\section{Results}\label{sec:results}

\smallskip \noindent {\bf End-to-End QA.}
For evaluation, we apply the relation prediction and entity detection models on each test question,
yielding a structured query $q=\{entityText\textrm{: }t_e, relation\textrm{: }r\}$ (Section~\ref{sec:approach}a). 
Entity linking gives us a list of candidate entity nodes (Section~\ref{sec:approach}b). 
For each candidate entity $e_{\textrm{cand}}$,
we can limit our relation choices to the set of unique relation types that some candidate entity $e_{\textrm{cand}}$ is associated with. 
This helps eliminate the artificial ambiguity due to overlapping relation types as well as the spurious ambiguity due to redundancies in 
a typical knowledge base. Even though there are 1837 relation types in Freebase, the number of relation types that we need to 
consider per question (on average) drops to 36.
The highest-scored answer node is selected by finding the highest scored entity node $e$ that has an outward edge of type $r$ 
(Section~\ref{sec:approach}c). 
We follow Bordes et al.~\shortcite{Bordes:2015aa} in comparing the predicted entity-relation pair to the ground truth. 
A question is counted as correct if and only if the entity we select (i.e., $e$) and the relation we predict (i.e, $r$)
match the ground truth.

Table~\ref{tab:end2end} summarizes end-to-end experimental results. We use the best models based on validation set accuracy
and compare it to three prior approaches:\ a specialized network architecture that explicitly memorizes facts~\cite{Bordes:2015aa},
a network that learns how to convolve sequence of characters in the question~\cite{Golub:2016aa}, and a complex network
with attention mechanisms to learn most important parts of the question~\cite{Yin:2016aa}.
Our approach outperforms the state of the art in accuracy (i.e., precision at top 1) by 11.9 points (15.6\% relative).


\begin{table}[hbt]
\centering
\begin{tabular}{|c|c|c|}
\hline
\textbf{Model} & \textbf{P@1} \\ \hline
Memory Network~\shortcite{Bordes:2015aa} & 63.9         \\ \hline
Char-level CNN~\shortcite{Golub:2016aa} & 70.9         \\ \hline
Attentive max-pooling~\shortcite{Yin:2016aa} & 76.4         \\ \hline\hline
RNN-QA (best models) & \textbf{88.3}  \\ \hline\hline
naive ED & 58.9         \\ \hline
naive RP & 4.1         \\ \hline
naive ED and RP & 3.7 \\ \hline
\end{tabular}
\caption{Top-1 accuracy on test portion of SimpleQuestions. Ablation study on last three rows.
}
\label{tab:end2end}
\end{table}

Last three rows quantify the impact of each component via an
\emph{\textbf{ablation study}}, in which we replace either entity detection (ED) or relation prediction (RP) models 
with a naive baseline:\ (i) we assign the relation that appears most frequently
in training data (i.e., {\small {\tt bornOn}}), and/or (ii) we tag the entire question as an entity (and then perform the $n$-gram
entity linking). Results confirm that RP is absolutely critical, since both datasets include a diverse and well-balanced
set of relation types. When we applied the naive ED baseline, our results drop significantly, but they are still 
comparable to prior results. Given that most prior work do not use the network to detect entities, we can deduce  
that our RNN-based entity detection is the reason our approach performs so well.



\noindent {\bf Error Analysis.}
 In order to better understand the weaknesses of our approach, we performed a \emph{blame analysis}:\ Among 2537 errors in 
 the test set, 15\% can be blamed on entity detection --- the relation type was correctly predicted, but the detected entity did not 
match the ground truth. The reverse is true for 48\% cases.\footnote{In remaining 37\% incorrect answers, both models fail, 
so the blame is shared.} We manually labeled a sample of 50 instances from each blame scenario. When entity
detection is to blame, 20\% was due to spelling inconsistencies between question and KB, which can be resolved with better 
text normalization during indexing (e.g., ``la kings'' refers to ``Los Angeles Kings''). We found 16\% of the detected entities to be correct, 
even though it was not the same as the ground truth (e.g., either ``New York'' or ``New York City'' is correct in ``what can do in 
new york?''); 18\% are inherently ambiguous and need clarification (e.g., ``where bin laden got killed?'' might mean ``Osama'' or ``Salem'').
When blame is on relation prediction, we found that the predicted relation is reasonable (albeit different than ground truth)
29\% of the time (e.g., ``what was nikola tesla known for'' can be classified as {\small {\tt profession}} or {\small {\tt notable\_for}}). 


%



\vspace{0.2cm}
\noindent {\bf RNN-QA in Practice.}
In addition to matching the state of the art in effectiveness, we also claimed that our simple architecture provides an efficient and modular 
solution. We demonstrate this by applying our model (without any modifications) to
the entertainment domain and deploying it to the Comcast X1 platform serving millions of customers every day.
Training data was generated synthetically based on an internal entertainment KB. For evaluation, 295 unique question-answer pairs 
were randomly sampled from real usage logs of the platform. 

We can draw two important conclusions from Table~\ref{tab:realworld}:\ First of all, we find that almost all of the user-generated 
natural-language questions (278/295$\sim$95\%) are first-order questions, supporting the significance of first-order QA as a task. 
Second, we show that even if we simply use an open-sourced deep learning toolkit (\url{keras.io}) for implementation and 
limit the computational resources to 2 CPU cores per thread, RNN-QA answers 75\% of questions correctly with very reasonable 
latency. 

\begin{table}[hbt]
\centering
\begin{tabular}{c|c|}
\textbf{Error} & \textbf{Count} \\ \hline
Correct 	      							&   220        \\ \hline
Incorrect entity			                                  &     16        \\ \hline
Incorrect relation			                          &     42        \\ \hline
Not first-order question          			&     17        \\ \hline\hline
Total Latency								&    76$\pm$16 ms
\end{tabular}
\caption{Evaluation of RNN-QA on real questions from X platform.}
\label{tab:realworld}
\end{table}

\section{Conclusions and Future work} 
We described a simple yet effective approach for QA, focusing primarily on first-order factual questions.
Although we understand the benefit of exploring task-agnostic approaches that aim to capture semantics
in a more general way (e.g., \cite{Kumar:2015aa}), it is also important to acknowledge that there is no ``one-size-fits-all''
solution as of yet. 

One of the main novelties of our work is to decompose the task into two subproblems, entity detection and relation prediction, and 
provide solutions for each in the form of a RNN. In both cases, we have found that bidirectional networks are beneficial, and that
two layers are sufficiently deep to balance the model's ability to fit versus its ability to generalize.

While an ablation study revealed the importance of both entity detection and relation prediction, we are hoping 
to further study the degree of which improvements in either component affect QA accuracy.
Drop-out was tuned to 10\% based on validation accuracy. While we have not implemented attention directly on our model, we
can compare our results side by side on the same benchmark task against prior work with complex attention mechanisms (e.g., \cite{Yin:2016aa}).
Given the proven strength of attention mechanisms, we were surprised to find our simple approach to be clearly superior on 
SimpleQuestions. 

Even though deep learning has opened the potential for more generic solutions, we believe that taking advantage 
of problem structure yields a more accurate and efficient solution. 
While first-order QA might seem like a solved problem, there is clearly still room for improvement. By revealing that 95\% of real use cases
fit into this paradigm, we hope to convince the reader that this is a valuable problem that requires more \emph{attention}.

\bibliographystyle{emnlp_natbib}
\bibliography{qa}

\begin{thebibliography}{28}
\expandafter\ifx\csname natexlab\endcsname\relax\def\natexlab#1{#1}\fi

\bibitem[{Berant et~al.(2013)Berant, Chou, Frostig, and Liang}]{Berant:2013aa}
Jonathan Berant, Andrew Chou, Roy Frostig, and Percy Liang. 2013.
\newblock \href {http://aclweb.org/anthology/D/D13/D13-1160.pdf} {Semantic
  parsing on freebase from question-answer pairs}.
\newblock In \emph{Proceedings of the 2013 Conference on Empirical Methods in
  Natural Language Processing, {EMNLP} 2013, 18-21 October 2013, Grand Hyatt
  Seattle, Seattle, Washington, USA, {A} meeting of SIGDAT, a Special Interest
  Group of the {ACL}}, pages 1533--1544. {ACL}.

\bibitem[{Bilotti et~al.(2010)Bilotti, Elsas, Carbonell, and
  Nyberg}]{Bilotti:2010aa}
Matthew~W Bilotti, Jonathan Elsas, Jaime Carbonell, and Eric Nyberg. 2010.
\newblock \href {https://doi.org/10.1145/1871437.1871498} {{Rank Learning for
  Factoid Question Answering with Linguistic and Semantic Constraints}}.
\newblock In \emph{Proceedings of the 19th ACM International Conference on
  Information and Knowledge Management}, CIKM '10, pages 459--468, New York,
  NY, USA. ACM.

\bibitem[{Bilotti et~al.(2007)Bilotti, Ogilvie, Callan, and
  Nyberg}]{Bilotti:2007aa}
Matthew~W Bilotti, Paul Ogilvie, Jamie Callan, and Eric Nyberg. 2007.
\newblock \href {https://doi.org/10.1145/1277741.1277802} {{Structured
  Retrieval for Question Answering}}.
\newblock In \emph{Proceedings of the 30th Annual International ACM SIGIR
  Conference on Research and Development in Information Retrieval}, SIGIR '07,
  pages 351--358, New York, NY, USA. ACM.

\bibitem[{Bordes et~al.(2014)Bordes, Chopra, and Weston}]{Bordes:2014aa}
Antoine Bordes, Sumit Chopra, and Jason Weston. 2014.
\newblock Question answering with subgraph embeddings.
\newblock \emph{arXiv preprint arXiv:1406.3676}.

\bibitem[{Bordes et~al.(2015)Bordes, Usunier, Chopra, and
  Weston}]{Bordes:2015aa}
Antoine Bordes, Nicolas Usunier, Sumit Chopra, and Jason Weston. 2015.
\newblock Large-scale simple question answering with memory networks.
\newblock \emph{CoRR}, abs/1506.02075.

\bibitem[{Buscaldi et~al.(2010)Buscaldi, Rosso, G\'{o}mez-Soriano, and
  Sanchis}]{Buscaldi:2010aa}
Davide Buscaldi, Paolo Rosso, Jos\'{e}~Manuel G\'{o}mez-Soriano, and Emilio
  Sanchis. 2010.
\newblock \href {https://doi.org/10.1007/s10844-009-0082-y} {{Answering
  Questions with an N-gram Based Passage Retrieval Engine}}.
\newblock \emph{J. Intell. Inf. Syst.}, 34(2):113--134.

\bibitem[{Chung et~al.(2014)Chung, G{\"{u}}l{\c{c}}ehre, Cho, and
  Bengio}]{Chung:2014aa}
Junyoung Chung, {\c{C}}aglar G{\"{u}}l{\c{c}}ehre, KyungHyun Cho, and Yoshua
  Bengio. 2014.
\newblock \href {http://arxiv.org/abs/1412.3555} {Empirical evaluation of gated
  recurrent neural networks on sequence modeling}.
\newblock \emph{CoRR}, abs/1412.3555.

\bibitem[{Collobert et~al.(2011)Collobert, Weston, Bottou, Karlen, Kavukcuoglu,
  and Kuksa}]{Collobert:2011aa}
Ronan Collobert, Jason Weston, L{\'e}on Bottou, Michael Karlen, Koray
  Kavukcuoglu, and Pavel Kuksa. 2011.
\newblock \href {http://dl.acm.org/citation.cfm?id=1953048.2078186} {Natural
  language processing (almost) from scratch}.
\newblock \emph{J. Mach. Learn. Res.}, 12:2493--2537.

\bibitem[{Dong et~al.(2015)Dong, Wei, Zhou, and Xu}]{Dong:2015ab}
Li~Dong, Furu Wei, Ming Zhou, and Ke~Xu. 2015.
\newblock \href {http://aclweb.org/anthology/P/P15/P15-1026.pdf} {Question
  answering over freebase with multi-column convolutional neural networks}.
\newblock In \emph{Proceedings of the 53rd Annual Meeting of the Association
  for Computational Linguistics and the 7th International Joint Conference on
  Natural Language Processing of the Asian Federation of Natural Language
  Processing, {ACL} 2015, July 26-31, 2015, Beijing, China, Volume 1: Long
  Papers}, pages 260--269. The Association for Computer Linguistics.

\bibitem[{Elman(1990)}]{Elman:1990aa}
Jeffrey~L Elman. 1990.
\newblock Finding structure in time.
\newblock \emph{Cognitive science}, 14(2):179--211.

\bibitem[{Golub and He(2016)}]{Golub:2016aa}
David Golub and Xiaodong He. 2016.
\newblock Character-level question answering with attention.
\newblock \emph{arXiv preprint arXiv:1604.00727}.

\bibitem[{Graves(2008)}]{Graves:2008aa}
Alex Graves. 2008.
\newblock \href {http://d-nb.info/99115827X} {\emph{Supervised sequence
  labelling with recurrent neural networks}}.
\newblock Ph.D. thesis, Technical University Munich.

\bibitem[{Hammerton(2003)}]{Hammerton:2003aa}
James Hammerton. 2003.
\newblock Named entity recognition with long short-term memory.
\newblock In \emph{Proceedings of the seventh conference on Natural language
  learning at HLT-NAACL 2003-Volume 4}, pages 172--175. Association for
  Computational Linguistics.

\bibitem[{He et~al.(2015)He, Gimpel, and Lin}]{He:2015aa}
Hua He, Kevin Gimpel, and Jimmy Lin. 2015.
\newblock \href {http://aclweb.org/anthology/D/D15/D15-1181.pdf}
  {Multi-perspective sentence similarity modeling with convolutional neural
  networks}.
\newblock In \emph{Proceedings of the 2015 Conference on Empirical Methods in
  Natural Language Processing, {EMNLP} 2015, Lisbon, Portugal, September 17-21,
  2015}, pages 1576--1586. The Association for Computational Linguistics.

\bibitem[{Hsu et~al.(2016)Hsu, Zhang, and Glass}]{Hsu:2016aa}
Wei{-}Ning Hsu, Yu~Zhang, and James~R. Glass. 2016.
\newblock \href {http://arxiv.org/abs/1603.07044} {Recurrent neural network
  encoder with attention for community question answering}.
\newblock \emph{CoRR}, abs/1603.07044.

\bibitem[{Huang et~al.(2015)Huang, Xu, and Yu}]{Huang:2015aa}
Zhiheng Huang, Wei Xu, and Kai Yu. 2015.
\newblock Bidirectional lstm-crf models for sequence tagging.
\newblock \emph{arXiv preprint arXiv:1508.01991}.

\bibitem[{Kumar et~al.(2015)Kumar, Irsoy, Su, Bradbury, English, Pierce,
  Ondruska, Gulrajani, and Socher}]{Kumar:2015aa}
Ankit Kumar, Ozan Irsoy, Jonathan Su, James Bradbury, Robert English, Brian
  Pierce, Peter Ondruska, Ishaan Gulrajani, and Richard Socher. 2015.
\newblock \href {http://arxiv.org/abs/1506.07285} {Ask me anything: Dynamic
  memory networks for natural language processing}.
\newblock \emph{CoRR}, abs/1506.07285.

\bibitem[{Lu et~al.(2015)Lu, Li, and Xu}]{Lu:2015aa}
Zefu Lu, Lei Li, and Wei Xu. 2015.
\newblock Twisted recurrent network for named entity recognition.
\newblock In \emph{Bay Area Machine Learning Symposium}.

\bibitem[{Mikolov et~al.(2013)Mikolov, Sutskever, Chen, Corrado, and
  Dean}]{Mikolov:2013aa}
Tomas Mikolov, Ilya Sutskever, Kai Chen, Greg~S Corrado, and Jeff Dean. 2013.
\newblock Distributed representations of words and phrases and their
  compositionality.
\newblock In \emph{Advances in neural information processing systems}, pages
  3111--3119.

\bibitem[{Mueller and Thyagarajan(2016)}]{Mueller:2016aa}
Jonas Mueller and Aditya Thyagarajan. 2016.
\newblock \href
  {http://www.aaai.org/ocs/index.php/AAAI/AAAI16/paper/view/12195} {Siamese
  recurrent architectures for learning sentence similarity}.
\newblock In \emph{Proceedings of the Thirtieth {AAAI} Conference on Artificial
  Intelligence, February 12-17, 2016, Phoenix, Arizona, {USA.}}, pages
  2786--2792. {AAAI} Press.

\bibitem[{Pennington et~al.(2014)Pennington, Socher, and
  Manning}]{Pennington:2014aa}
Jeffrey Pennington, Richard Socher, and Christopher~D Manning. 2014.
\newblock Glove: Global vectors for word representation.
\newblock In \emph{EMNLP}, volume~14, pages 1532--1543.

\bibitem[{Rao et~al.(2016)Rao, He, and Lin}]{Rao:2016aa}
Jinfeng Rao, Hua He, and Jimmy Lin. 2016.
\newblock Noise-contrastive estimation for answer selection with deep neural
  networks.
\newblock In \emph{Proceedings of the 25th ACM International on Conference on
  Information and Knowledge Management}, pages 1913--1916. ACM.

\bibitem[{Reddy et~al.(2014)Reddy, Lapata, and Steedman}]{Reddy:2014aa}
Siva Reddy, Mirella Lapata, and Mark Steedman. 2014.
\newblock \href
  {https://tacl2013.cs.columbia.edu/ojs/index.php/tacl/article/view/398}
  {Large-scale semantic parsing without question-answer pairs}.
\newblock \emph{{TACL}}, 2:377--392.

\bibitem[{Wang et~al.(2015)Wang, Qian, Soong, He, and Zhao}]{Wang:2015aa}
Peilu Wang, Yao Qian, Frank~K Soong, Lei He, and Hai Zhao. 2015.
\newblock Part-of-speech tagging with bidirectional long short-term memory
  recurrent neural network.
\newblock \emph{arXiv preprint arXiv:1510.06168}.

\bibitem[{Yang et~al.(2016)Yang, Ai, Guo, and Croft}]{Yang:2016aa}
Liu Yang, Qingyao Ai, Jiafeng Guo, and W~Bruce Croft. 2016.
\newblock anmm: Ranking short answer texts with attention-based neural matching
  model.
\newblock In \emph{Proceedings of the 25th ACM International on Conference on
  Information and Knowledge Management}, pages 287--296. ACM.

\bibitem[{Yang et~al.(2014)Yang, Duan, Zhou, and Rim}]{Yang:2014aa}
Min-Chul Yang, Nan Duan, Ming Zhou, and Hae-Chang Rim. 2014.
\newblock Joint relational embeddings for knowledge-based question answering.
\newblock In \emph{EMNLP}, pages 645--650.

\bibitem[{Yin et~al.(2016)Yin, Yu, Xiang, Zhou, and Sch{\"u}tze}]{Yin:2016aa}
Wenpeng Yin, Mo~Yu, Bing Xiang, Bowen Zhou, and Hinrich Sch{\"u}tze. 2016.
\newblock Simple question answering by attentive convolutional neural network.
\newblock \emph{arXiv preprint arXiv:1606.03391}.

\bibitem[{Zhang and Lee(2003)}]{Zhang:2003aa}
Dell Zhang and Wee~Sun Lee. 2003.
\newblock \href {https://doi.org/10.1145/860435.860443} {{Question
  Classification Using Support Vector Machines}}.
\newblock In \emph{Proceedings of the 26th Annual International ACM SIGIR
  Conference on Research and Development in Informaion Retrieval}, SIGIR '03,
  pages 26--32, New York, NY, USA. ACM.

\end{thebibliography}

\end{document}